\documentclass[runningheads]{llncs}

\usepackage{tabularx}
\usepackage{subfigure}
\usepackage[mathfrak]{mathpi}

\usepackage{times}
\usepackage{graphicx}
\usepackage{array}
\usepackage{amsmath}
\usepackage{amssymb}
\usepackage{algorithm}
\usepackage{algorithmic}
\usepackage{url}
\usepackage{setspace}
\usepackage{multirow}

\graphicspath{{imgs/}}

\newcolumntype{V}{>{$\vcenter\bgroup\hbox\bgroup}c<{\egroup\egroup$}}

\begin{document}

\pagestyle{headings}

\mainmatter

\title{Multiple-Object Tracking in Cluttered and Crowded Public Spaces}
\titlerunning{Lecture Notes in Computer Science}
\author{Rhys Martin \and Ognjen Arandjelovi\'c}
\institute{University of Cambridge\\Department of Engineering\\Cambridge CB2 1TQ, UK}

\maketitle

\begin{abstract}
This paper addresses the problem of tracking moving objects of
variable appearance in challenging scenes rich with features and
texture. Reliable tracking is of pivotal importance in surveillance
applications. It is made particularly difficult by the nature of
objects encountered in such scenes: these too change in appearance
and scale, and are often articulated (e.g.\ humans). We propose a
method which uses fast motion detection and segmentation as a
constraint for both building appearance models and their robust
propagation (matching) in time. The appearance model is based on
sets of local appearances automatically clustered using
spatio-kinetic similarity, and is updated with each new appearance
seen. This integration of all seen appearances of a tracked object
makes it extremely resilient to errors caused by occlusion and the
lack of permanence of due to low data quality, appearance change or
background clutter. These theoretical strengths of our algorithm are
empirically demonstrated on two hour long video footage of a busy
city marketplace.
\end{abstract}

\section{Introduction}\label{s:intro}
In recent years the question of security in public spaces has been
attracting an increasing amount of attention. While the number of
surveillance cameras has steadily increased so have the problems
associated with the way vast amounts of collected data are used. The
inspection of recordings by humans is laborious and slow, and as a
result most surveillance footage is used not preventatively but
rather \textit{post hoc}. Work on automating this process by means
of computer vision algorithms has the potential to be of great
public benefit and could radically change how surveillance is
conducted.

Most objects of interest in surveillance footage move at some point
in time. Tracking them reliably is a difficult but necessary step
that needs to be performed before any inference at a higher level of
abstraction is done. This is the problem we address in this paper.

\subsection{Problem Difficulties}\label{ss:difficulties}
\begin{figure*}[t]
  \centering
  \small
  \begin{tabular}{VV}
    \includegraphics[height=0.3\textwidth]{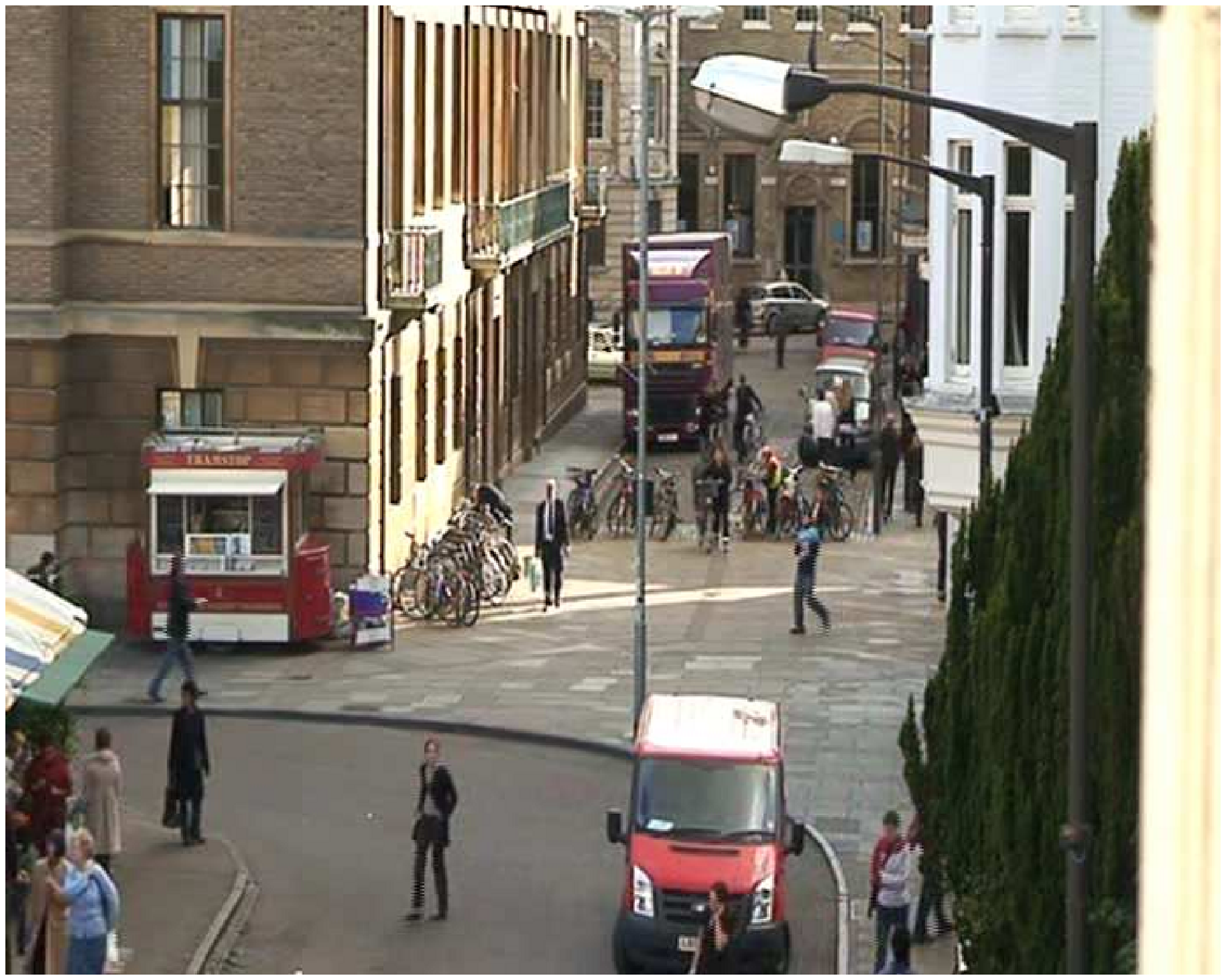} &
    \includegraphics[height=0.3\textwidth]{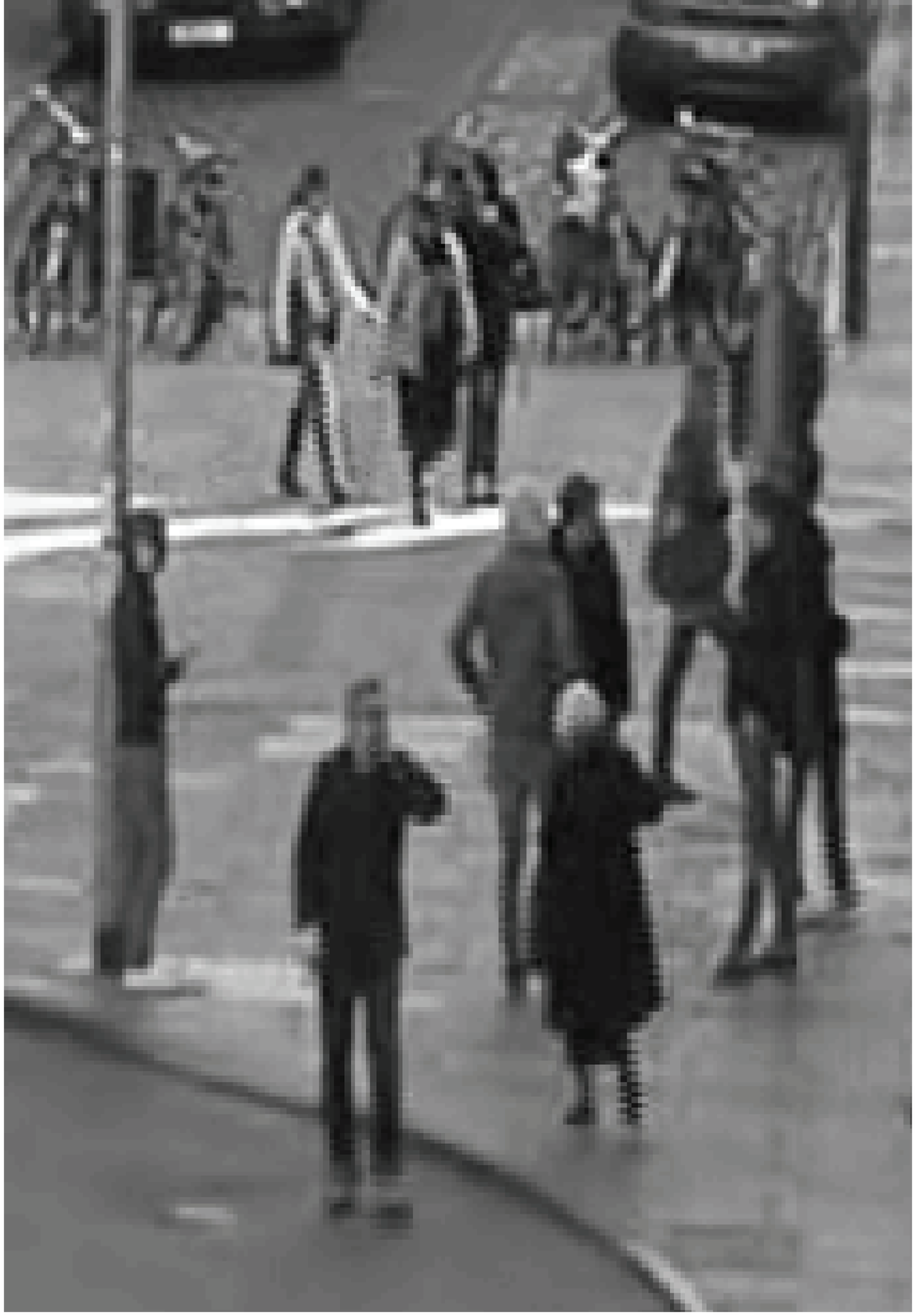}\\
    & \\
    (a) Original & (b) Magnification \\&\\
    \includegraphics[height=0.3\textwidth]{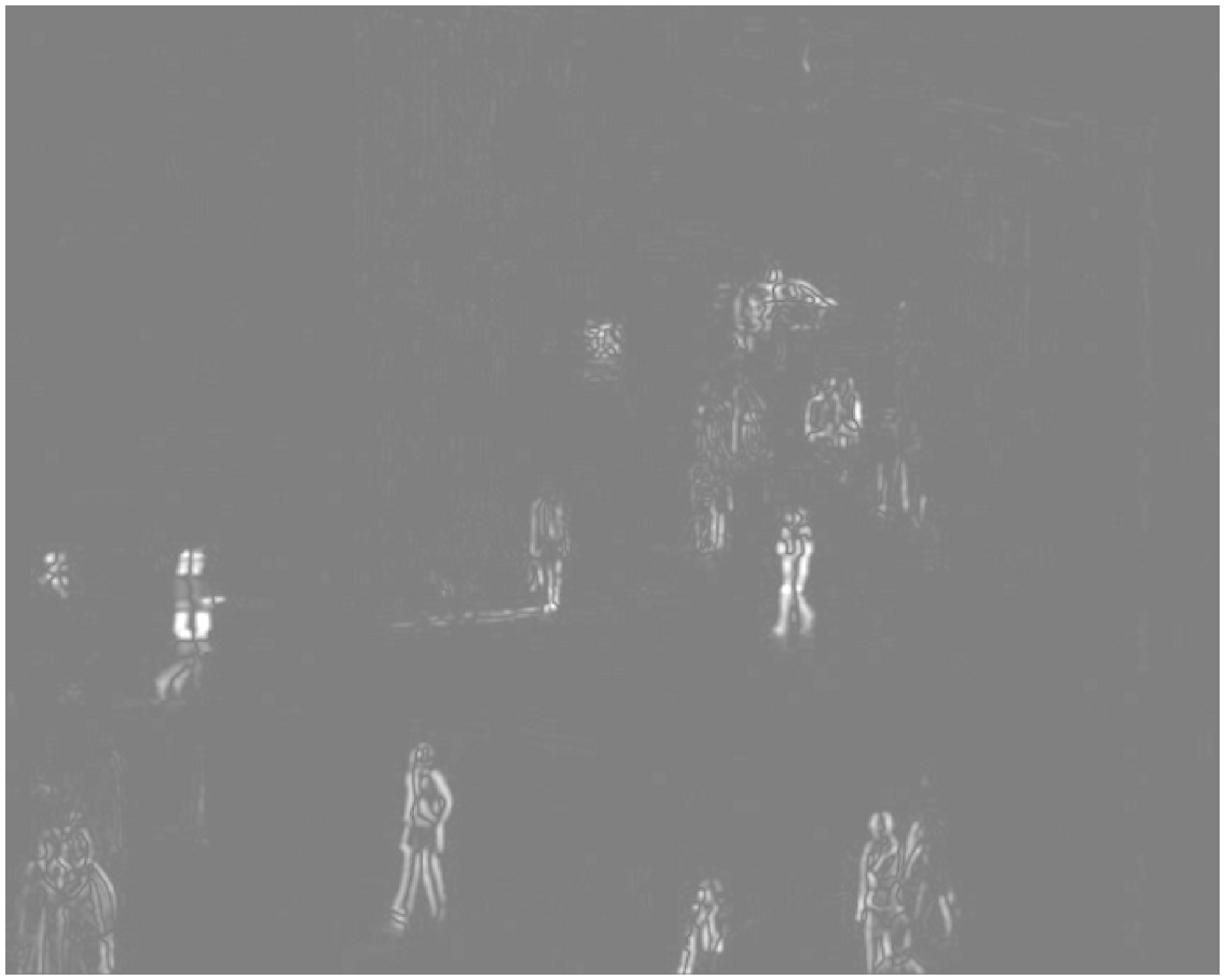} &
    \includegraphics[height=0.3\textwidth]{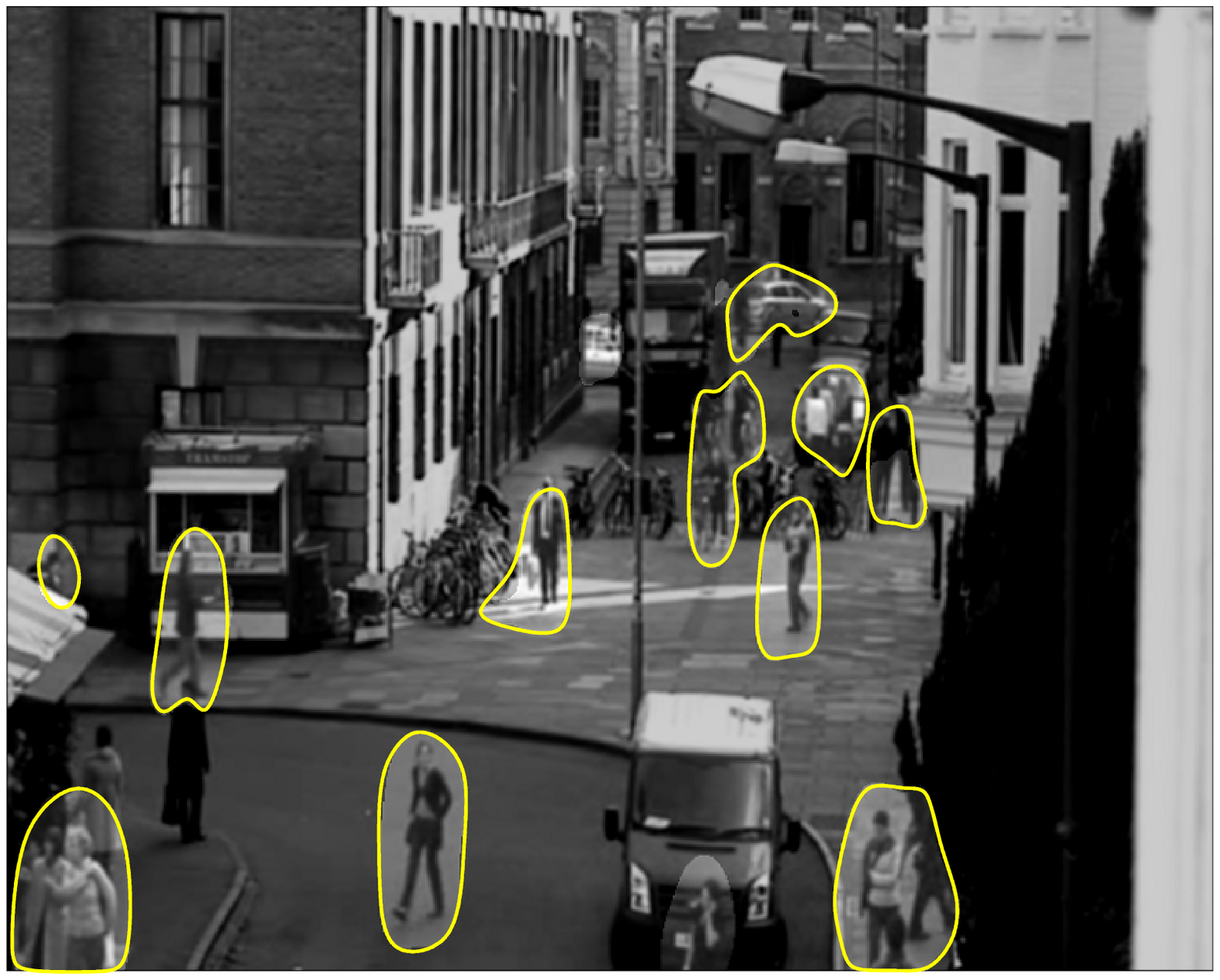}\\
    &\\
    (c) Difference & (d) Motion regions\\\vspace{10pt}
  \end{tabular}
  \caption{(a) A typical frame extracted from the video footage used in the evaluation of this paper,
            showing a busy city marketplace, (b) a magnified image region containing examples of occlusion
            of objects of interests, as well as a cluttered and feature rich background, (c) the difference between
            two successive frames and (d) the inferred motion regions.}
            \label{f:frame}
\end{figure*}

Public spaces are uncontrolled and extremely challenging
environments for computer vision-based inference. Not only is their
appearance rich in features, texture and motion -- as exemplified in
Figures~\ref{f:frame}~(a) and \ref{f:frame}~(b) -- but it is also
continuously exhibiting variation of both high and low frequency in
time: shopping windows change as stores open and close, shadows cast
by buildings and other landmarks move, delivery lorries get parked
intermittently etc. This is a major obstacle to methods based on
learning the appearance of the background, e.g.\
\cite{WrenAzarDarrPent1997,HariHarwDavi2000,ZhaoNeva2004,IsarMacC2001}.

Indeed, little related previous research addressed the exact problem
we consider in this paper, instead concentrating on simpler
recognition and tracking environments. A popular group of methods is
based on grouping low-level features, for example by detecting
common motion patterns \cite{RabaBelo2006,BrosCipo2006} or using
cascaded appearance classifiers \cite{Gavr2000,ViolJoneSnow2003}.
While these tend to perform well in uncrowded scenes, they have
difficulties coping with occlusions. This is particularly the case
with mutual occlusions, involving multiple tracked objects. Both
methods of Rabaud and Belongie \cite{RabaBelo2006}, and Brostow and
Cipolla\cite{BrosCipo2006}, share some similarity with the method
proposed in this paper, in that they consider the coherence of
feature motion. However, unlike our method, their approaches rely on
having long, reliable tracks of interest points. Our experiments
suggests that this is not a realistic assumption for uncontrolled
crowded scenes -- local features are difficult to detect reliably in
videos of the kind of quality which is found in practice. These
usually have poor resolution and are often compressed, making most
local features very short lived. This is complicated further by
frequent occlusion and object articulation.

In contrast, template-based methods which employ holistic appearance
struggle with the issue of variability in appearance of tracked
objects and their scale
\cite{TuSebaDoreKrah2008,ZhaoNevaLv2001,LiptFujiSapi1998,MattIshiBake2004},
and generally have high computational demands. Zhao and Nevatia
\cite{ZhaoNeva2004}, for example, employ more restrictive object and
scene models, in the form of human shape and ground plane
calibration and assume a bird's eye view of the scene. This approach
is thus fundamentally more restrictive than ours in several
important aspects. Additionally, unlike ours, models of this kind
struggle with the problem of initialization which is usually manual.
This is a major limitation in a practical application which involves
a great number of moving entities which uncontrollably enter and
leave the scene.



In summary, an algorithm successful at tracking moving entities in a
crowded public space, has to, on the one hand, learn a model
sufficiently persistent and discriminative to correctly track in the
presence of occlusion and distinguish between potentially similar
entities, yet flexible enough to allow for appearance changes due to
articulation, pose and scale change. In the next section we describe
the details of our approach at achieving this, followed by a section
in which its performance is illustrated on real-world footage of a
public square.

\section{Algorithm Details}
Our algorithm employs multiple (and in a sense complementary)
representations as a means of capturing a suitably strong model that
allows for reliable tracking and track continuity following partial
or full occlusion, while at the same time exhibiting sufficient
flexibility in changing appearance and computational efficiency. We
first give an overview of the approach, followed by a detailed
description of each of the steps.

\subsection{Overview}
The proposed method consists of an interlaced application of the
following key algorithmic elements:
\begin{itemize}
  \item Detection of motion regions
        (in all frames, across the entire frame area)
  \item Spatial grouping of motion regions 
  \item Interest point detection
        (within motion regions only)
  \item Appearance model building by spatio-kinetic clustering of interest points (newly detected ones only)
  \item Correspondence matching of feature clusters between successive frames
\end{itemize}
We build appearance models from bottom up, grouping local features
within motion regions into clusters, each cluster representing a
moving object, according to the coherence of their motion and taking
into account perspective effects of the scene. Permanence of
appearance models is achieved by retaining all features added to a
cluster even after their disappearance (which often happens, due to
occlusion, articulation, or image noise, for example). Robustness in
searching for feature and object correspondence between frames is
gained by using constraints derived from detected motion regions,
allowing us to account for occlusion or transiently common motion of
two objects.

\subsection{Detecting and Grouping Motion Regions}\label{ss:regions}
An important part of the proposed method lies in the use of motion
regions. These are used to dramatically improve computational
efficiency, reducing the image area which is processed further by
focusing only on its ``interesting'' parts, as well as to constrain
the feature correspondence search -- described in
Section~\ref{ss:tracking} -- which is crucial for reliable matching
of appearance models of moving entities between subsequent frames.

Let $I_t \in \mathbb{R}^{H\times W}$ be the frame (as a $H \times W$
pixel image) at the $t$-th time step in the input video. At each
time step, our algorithm performs simple motion detection by
pixel-wise subtraction of two frames $k$ steps apart:
\begin{align}
  \Delta I_{t}(x,y) = I_{t}(x,y) - I_{t-k}(x,y).
\end{align}
Typical output is illustrated in Figure~\ref{f:frame}~(c) which
shows rather noisy regions of appearance change. Note that many
locations which correspond to moving entities by coincidence do not
necessarily significantly change in appearance. To account for this,
we employ the observation that the objects of interest have some
expected spatial extent. Thus, we apply a linear smoothing operator
on the frame difference $\Delta I_{t}(x,y)$:
\begin{align}
  C_t(x,y) = \int_{u,v} \Delta I_t(x+u,y+v)~G(u,v,y)
\end{align}
where $G(u,v,y)$ is an \emph{adaptive} Gaussian filter.
Specifically, the variances of the axis-aligned kernel are made
dependent on the location of its application:
\begin{align}
  G\left(u,v,y~|~\,\sigma_u, \sigma_v\right) =
    \frac{1}{2\pi~\sigma_u~\sigma_v } \exp\Big\{ - 0.5~u^2 / \sigma_u(y) - 0.5~v^2 / \sigma_v(y)
    \Big\}.
\end{align}
The variation of $\sigma_u(y)$ and $\sigma_v(y)$ is dependent on the
scene perspective and the loose shape of the objects of interest. We
learn them in the form $\sigma_u(y) = c_1~y + c_2$ and $\sigma_v(y)
= c_3~y + c_2$. As our appearance model (described next) is
top-down, that is, initial hypotheses for coherently moving entities
are broken down, rather than connected up, we purposefully choose
relatively large $c_1$ and $c_3$ ($0.045$ and $0.25$, respectively).
The remaining constant is inferred through minimal user input: the
user is asked to select two pairs of points such that the points in
each pair are at the same distance from the camera and at the same
distance from each other, and that each pair is at a different
distance from the camera.

Finally, we threshold the result and find all connected components
consisting of positively classified pixels (those exceeding the
threshold) which we shall for brevity refer to as  motion regions.
On our data set, on average they occupy approximately 8\% of the
total frame area. Examples are shown in Figure~\ref{f:frame}~(d).

\subsection{Building Appearance Models using Spatio-Kinetic Clustering of Interest Points}\label{ss:model}
Having identified regions of interest in the scene, we extract
interest points in them as scale-space maxima \cite{Lowe2001}. While
motion regions are used to constrain their matching and clustering,
descriptors of local appearance at interest points are collectively
used to represent the appearance of tracked objects.

Each interest point's circular neighbourhood is represented by the
corresponding 128-dimensional SIFT descriptor \cite{Lowe2001}. These
are then grouped according to the likelihood that they belong to the
same object. Exploiting the observation that objects have limited
spatial extent, as well as that their constituent parts tend to move
coherently, we cluster features using both spatial and motion cues,
while accounting for the scene geometry.

\begin{figure}[htp]
  \centering
  \footnotesize
  \begin{tabular}{VVV}
    \includegraphics[width=0.25\textwidth]{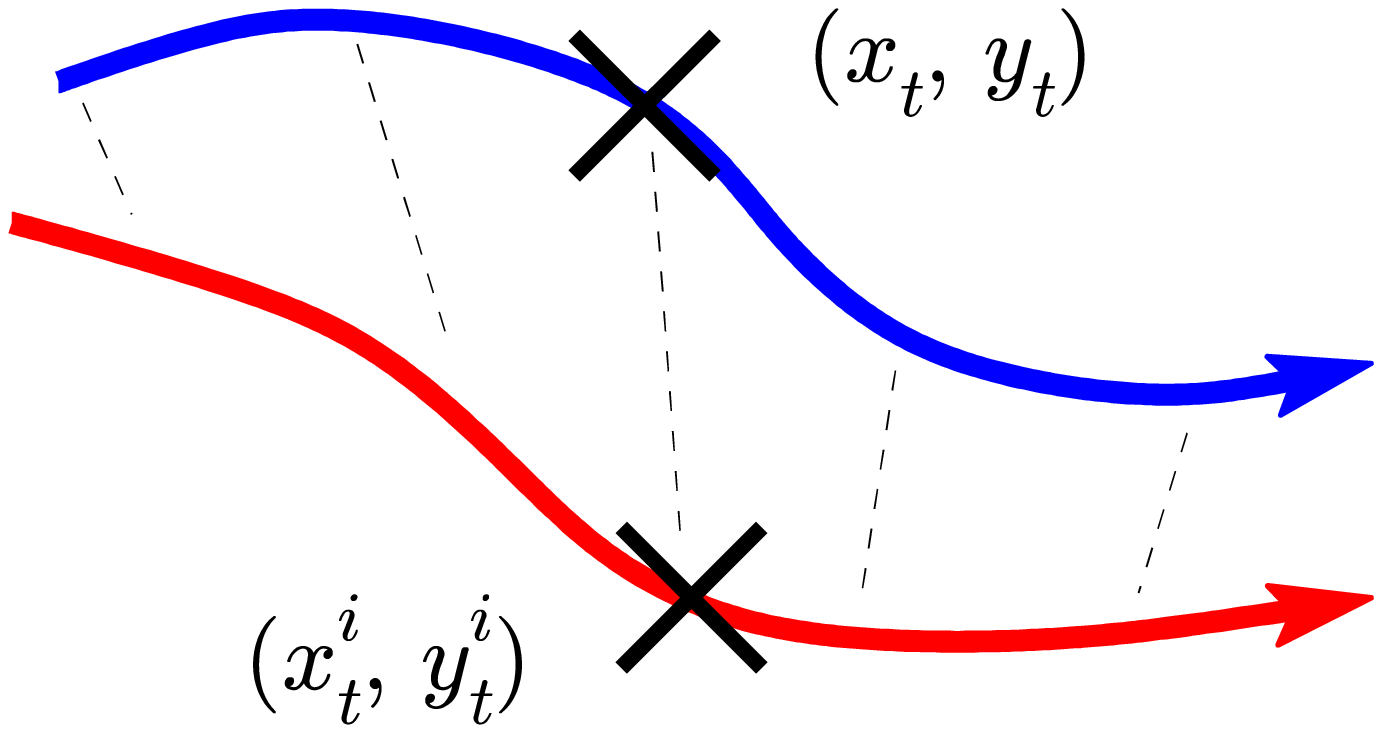}&
    \includegraphics[width=0.25\textwidth]{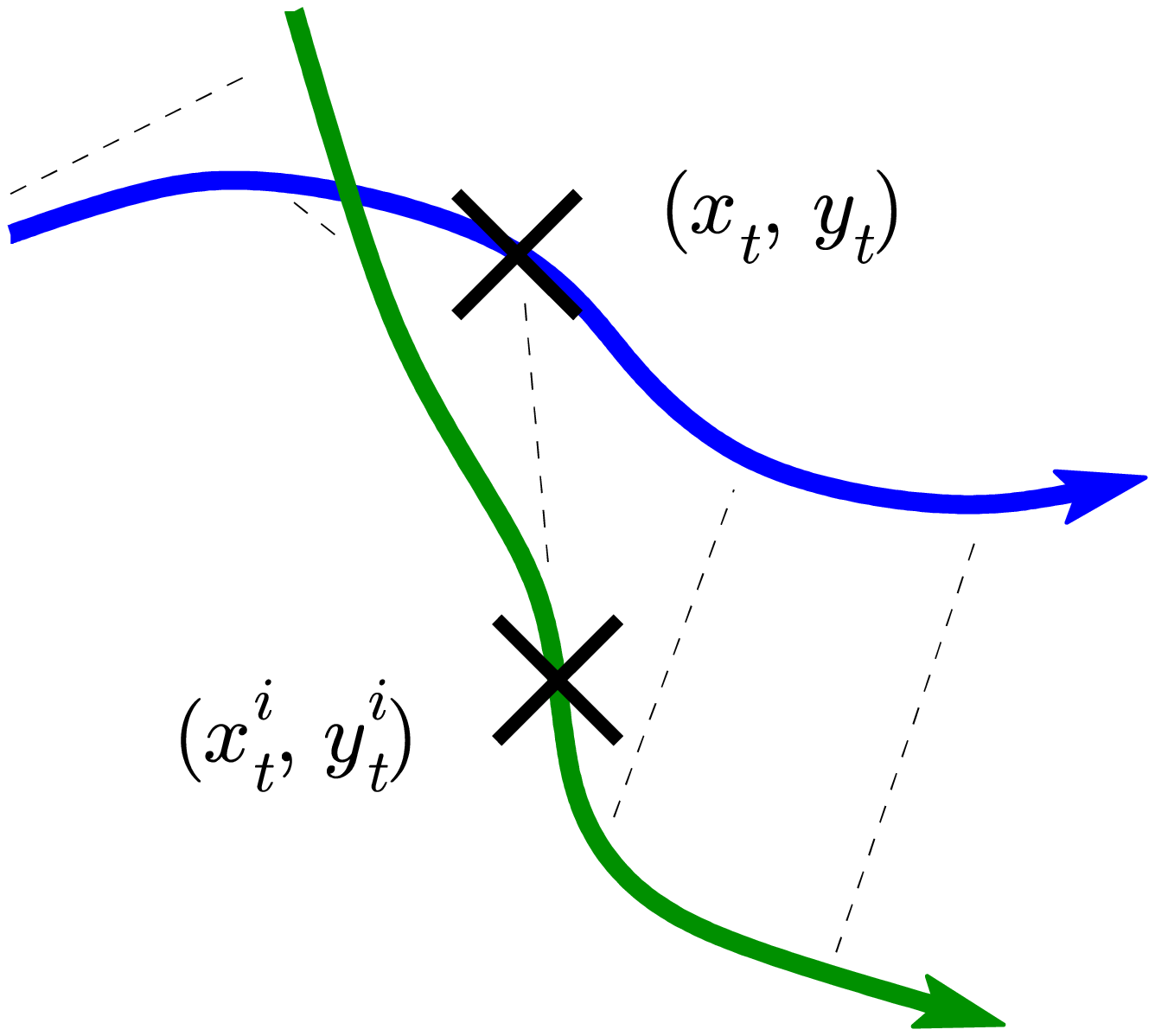}&
    \includegraphics[width=0.30\textwidth]{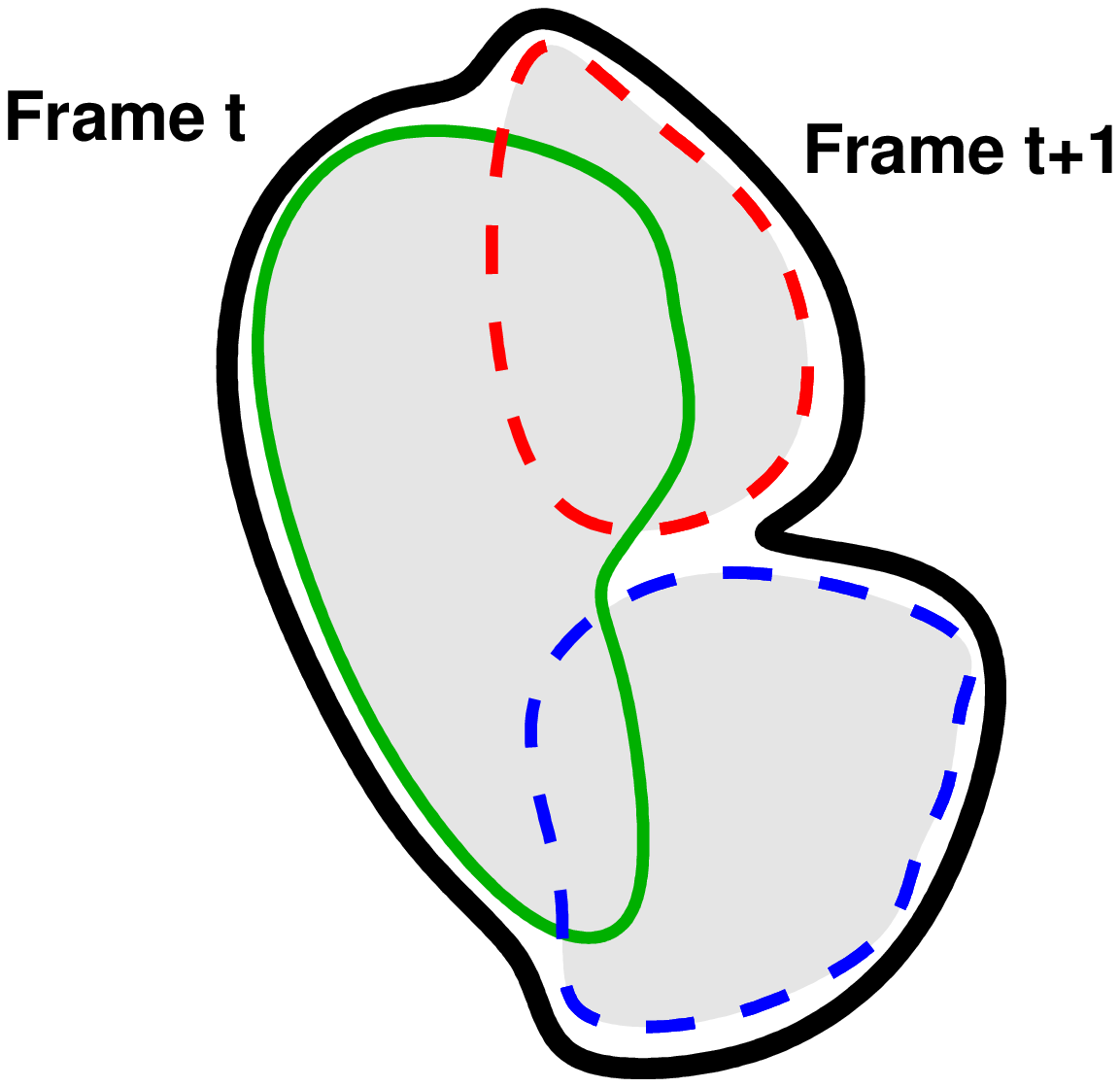}\\
    &&\\
    (a) & (b) & (c)\\
    &&\\
  \end{tabular}
  \caption{Kinetic (a) coherence and (b) discrepancy result in two features
            with spatial proximity getting assigned to respectively the same and
            different clusters. (c) A feature located in a specific motion region
            in frame $I_t$ is searched for in the subsequent frame $I_{t+1}$ in the
            area occupied by the initial motion region (green, solid) and all motion
            regions that intersect it in $I_{t+1}$ (blue and red, dashed).}
  \label{f:features}
\end{figure}

The spatial constraint is applied by virtue of hierarchial
clustering -- only the $K$ nearest neighbours of each interest point
are considered in trying to associate it with an existing cluster.
Using a limited velocity model, an interest point and its neighbour
are tracked $N$ frames forwards and backwards in time to extract the
corresponding motion trajectories. Let the motion of a tracked
interest point be described by a track of its location through time
$\{(x_t,y_t)\} = \{(x_{t_1},y_{t_1}), (x_{t_1+1},y_{t_1+1}), \ldots,
(x_{t_2},y_{t_2})$ and that of its $i$-th of $K$ nearest neighbours
$\{(x^i_t,y^i_t)\} = \{(x^i_{t_1},y^i_{t_1}),
(x^i_{t_1+1},y^i_{t_1+1}), \ldots, (x^i_{t_2},y^i_{t_2})$, where the
interval $[t_1,t_2]$ is determined by the features' maximal past and
future co-occurrence. The two interest points are associated with
the same appearance cluster -- a cluster being the current best
hypothesis of a single moving entity -- if they have not been
already associated with separate clusters and the motion incoherence
of the corresponding trajectories does not exceed a threshold
$t_{\text{coherence}}$:
\begin{align}
  \sum_{t=t_1}^{t_2} \left \| \frac{ (x_t,y_t) - (x^i_t,y^i_t)} { (y_t + y^i_t)~/~2 + c_2 } \right \|^2 -
  \left( \sum_{t=t_1}^{t_2} \left \| \frac{ (x_t,y_t) - (x^i_t,y^i_t)} { (y_t + y^i_t)~/~2 + c_2 } \right
  \| \right)^2 < t_{\text{coherence}}.
\label{e:coherence}
\end{align}
as conceptually illustrated in Figures~\ref{f:features}~(a) and
\ref{f:features}~(b). The coherence measure in
Equation~\ref{e:coherence} accounts for the previously learnt
perspective of the scene by inversely weighting the distance between
two features by their distance from the horizon. Note that we make
the implicit assumption that the vertical position of the camera is
significantly greater than the height of tracked objects (if this
assumption is invalidated, the denominators in
Equation~\ref{e:coherence} can reach a small value without the
objects being near the horizon).

The result of the described spatio-kinetic clustering is a set of
clusters per each motion region. These are associated with the
region only temporarily and it is not assumed that they correspond
to the same object (indeed, in most cases they do not due to
different motion characteristics).

\subsection{Model Propagation through Time}\label{ss:tracking}
Having learnt quasi-permanent appearance models of objects (as their
constituent features are being detected using the approach described
in Section~\ref{ss:model}), we turn our attention to the question of
tracking these through time.

Consider a cluster of features in a particular motion region and the
problem of localizing this cluster in the subsequent frame. We know
that the features which belong in it move coherently and we know
that this motion is limited in velocity. However, the corresponding
motion region may no longer exist: the features may have temporarily
ceased moving, or the objects which comprised a single motion region
may have parted (e.g.\ two people separating, or after temporary
occlusion), or it may have joined another (e.g.\ two people meeting
or occluding each other). To account for all of these possibilities,
each feature in searched for in the area occupied by the original
region it was detected in and all the regions in the subsequent
frame which intersect it, as shown in Figure~\ref{f:features}~(c).

\begin{figure}[htp]
  \centering
  \includegraphics[width=0.9\textwidth]{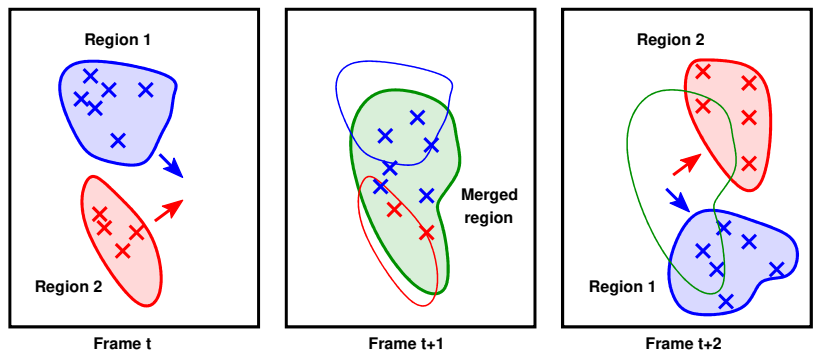}
  \caption{A conceptual illustration showing the robustness of our appearance
           model in coping with partial or full occlusion regardless of its duration.} \label{f:occ1}
\end{figure}

Consider an interest point with the appearance at the time step $t$
captured by the corresponding SIFT descriptor $\mathbf{d}_t \in
\mathbb{R}^{128}$. It is matched to that in the next frame $t+1$ and
within the search area, which has the most similar appearance,
$\mathbf{d}_{t+1}^k$, provided that their similarity exceeds a set
threshold according to the following criterion:
\begin{align}
  &\mathbf{d}_t~\xrightarrow{~match~}~\mathbf{d}_{t+1}^k
\end{align}
where
\begin{align}
  k=
  \begin{cases}
     \arg \min_i~\rho(i) & \rho(k) \leq t_{\text{feature}} \\
     \text{new feature}  & \rho(k) > t_{\text{feature}} \\
  \end{cases}
&&~\text{~and~}&
  &\rho(i) = \frac {{\mathbf{d}_t}^T~\mathbf{d}_{t+1}^i} {\| \mathbf{d}_t
  \|~\| \mathbf{d}_{t+1}^i \|}
\end{align}
Features from the same cluster which are localized within the same
motion region in the new frame are associated with it, much like
when the model is first built, as described in
Section~\ref{ss:model}. However, the cluster can also split when its
constituent features are localized in different motion regions
(e.g.\ when two people who walked together separate). Cluster
splitting is effected by splitting the appearance model and
associating each new cluster with the corresponding motion region.
On the other hand, notice that clusters are never joined even if
their motion regions merge, as illustrated in Figure~\ref{f:occ1}.
This is because it is the clusters themselves which represent the
best current estimate of individual moving objects in the scene,
whereas motion regions merely represent the image plane uncertainty
in temporal correspondence, caused by motion of independent
entities.


\subsection{Review of Handling of the Key Tracking Problems}
As a conclusion to the theoretical section of the paper, let us
consider how our algorithm copes with some of the most important
tracking scenarios.

\begin{itemize}
  \item \textbf{Case 1, Independently moving entity:} An appearance model is
  built from the appearances of clustered local features. As the cluster is tracked
  by matching those features which are at the time reliably matched,
  the model in the form of a set of appearance features (many of which are
  \emph{not} visible at any point in time) is constantly enriched as
  new appearances are observed.

  \item \textbf{Case 2, Coherently moving entities, which separate:} Motion
  incoherence after separation is used to infer separate entities,
  which are back-tracked to the time of common motion when the
  corresponding clusters (as their feature sets) are associated with
  the same moving region. Novel appearance, in the form of new local
  features is added to the correct appearance cluster using spatial
  constraints.

  \item \textbf{Case 3, Separately moving entities, which join in their motion:}
  This situation is handled in the same manner as that described previously as
  Case 2, but with tracking proceeding forwards, not backwards in time.

  \item \textbf{Case 4, Partial occlusion of a tracked entity:}
  The proposed appearance
  model in the form of a set of appearances of local features, is
  inherently robust to partial occlusion -- correct correspondence
  between clusters is achieved by matching reliably tracked, visible
  features.

  \item \textbf{Case 5, Full occlusion of a tracked entity:} When a
  tracked entity is occluded by another, both of their clusters are
  associated with the same motion region. This association
  continues until sufficient evidence for the occluded entity
  re-emerges and a new motion region is detected. At that point the
  overlap of regions of interest is used to correctly match appearance
  models, separating them and re-assigning feature clusters with the
  correct moving regions.
\end{itemize}
An empirical demonstration of these theoretical arguments is
presented next.

\section{Empirical Analysis}
To evaluate the effectiveness of the proposed method we acquired a
data set fitting the problem addressed in this paper and containing
all of the challenging aspects described in
Section~\ref{ss:difficulties}. Using a stationary camera placed on
top of a small building overlooking a busy city marketplace we
recorded a continuous video stream of duration 1h:59m:40s and having
the spatial resolution of $720 \times 576$ pixels. A typical frame
is shown in Figure~\ref{f:frame}~(a) while Figure~\ref{f:frame}~(b)
exemplifies some of the aforementioned difficulties on a magnified
subregion.

\begin{figure}[htp]
  \centering
  \footnotesize
  \begin{tabular}{VV}
    \rotatebox{-90}{\includegraphics[width=0.4\textwidth]{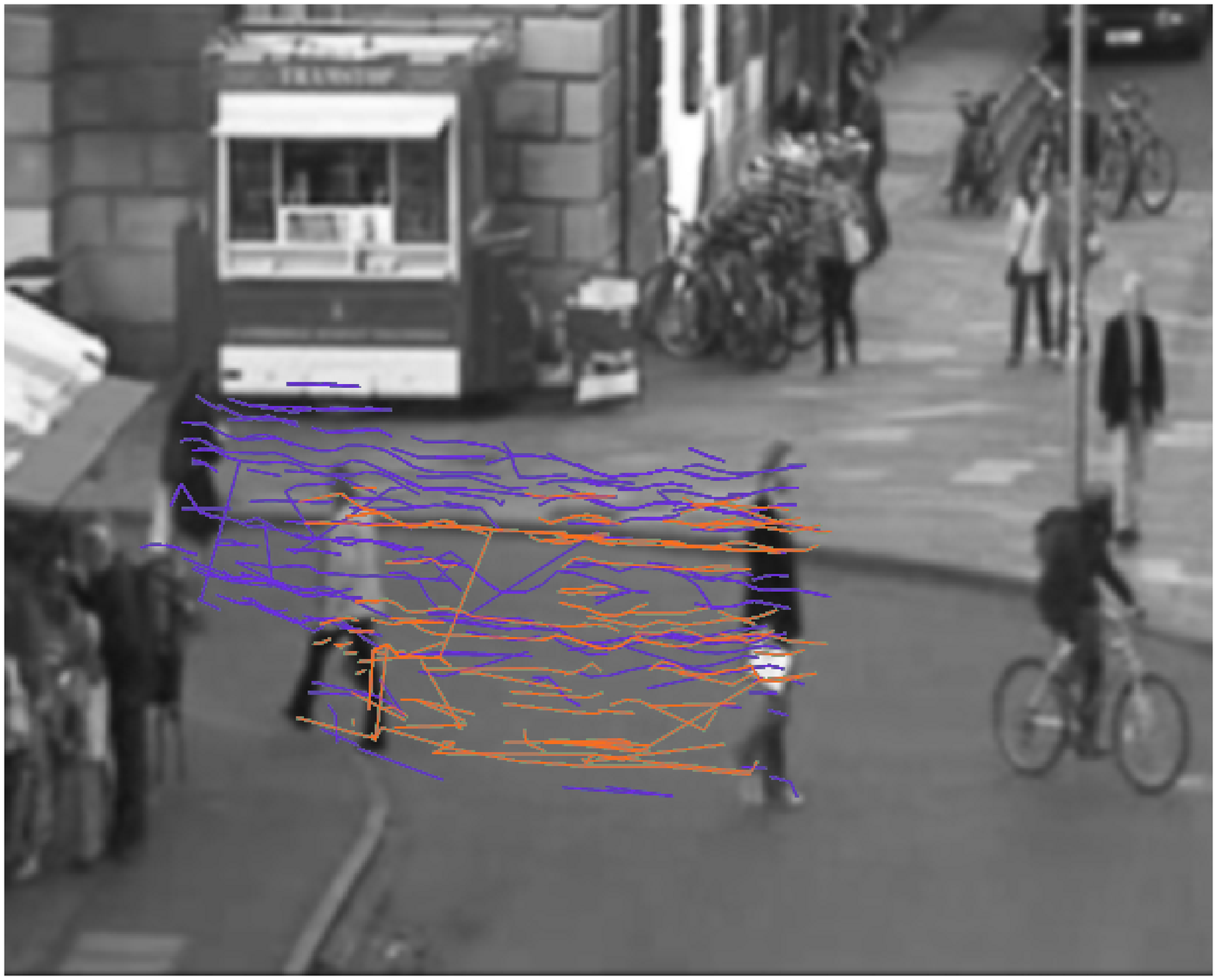}}
    &
    \rotatebox{-90}{\includegraphics[width=0.4\textwidth]{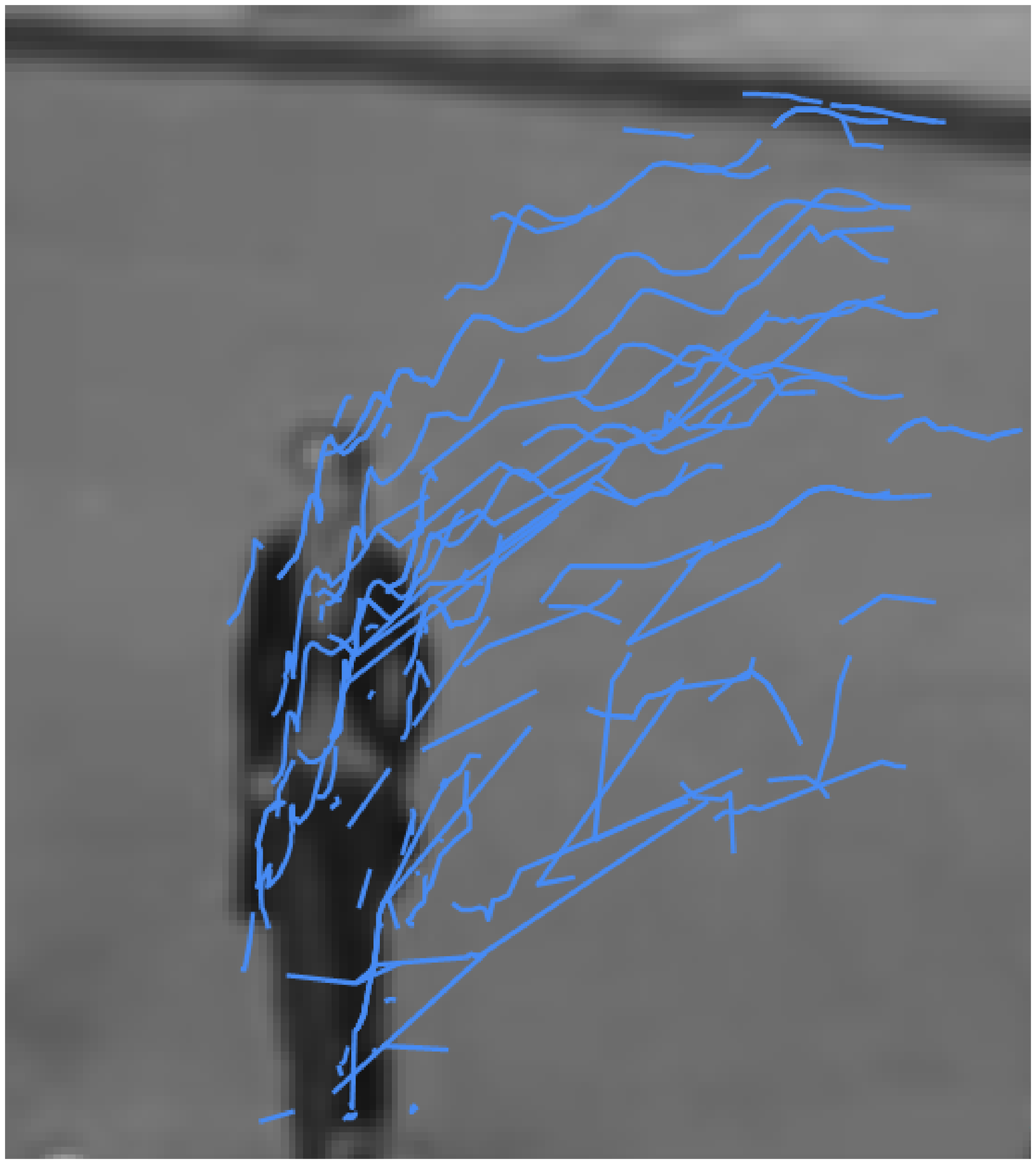}}\\&\\
    (a) & (b) \\&\\
  \end{tabular}\\
  \begin{tabular}{VV}
    \rotatebox{-90}{\includegraphics[width=0.4\textwidth]{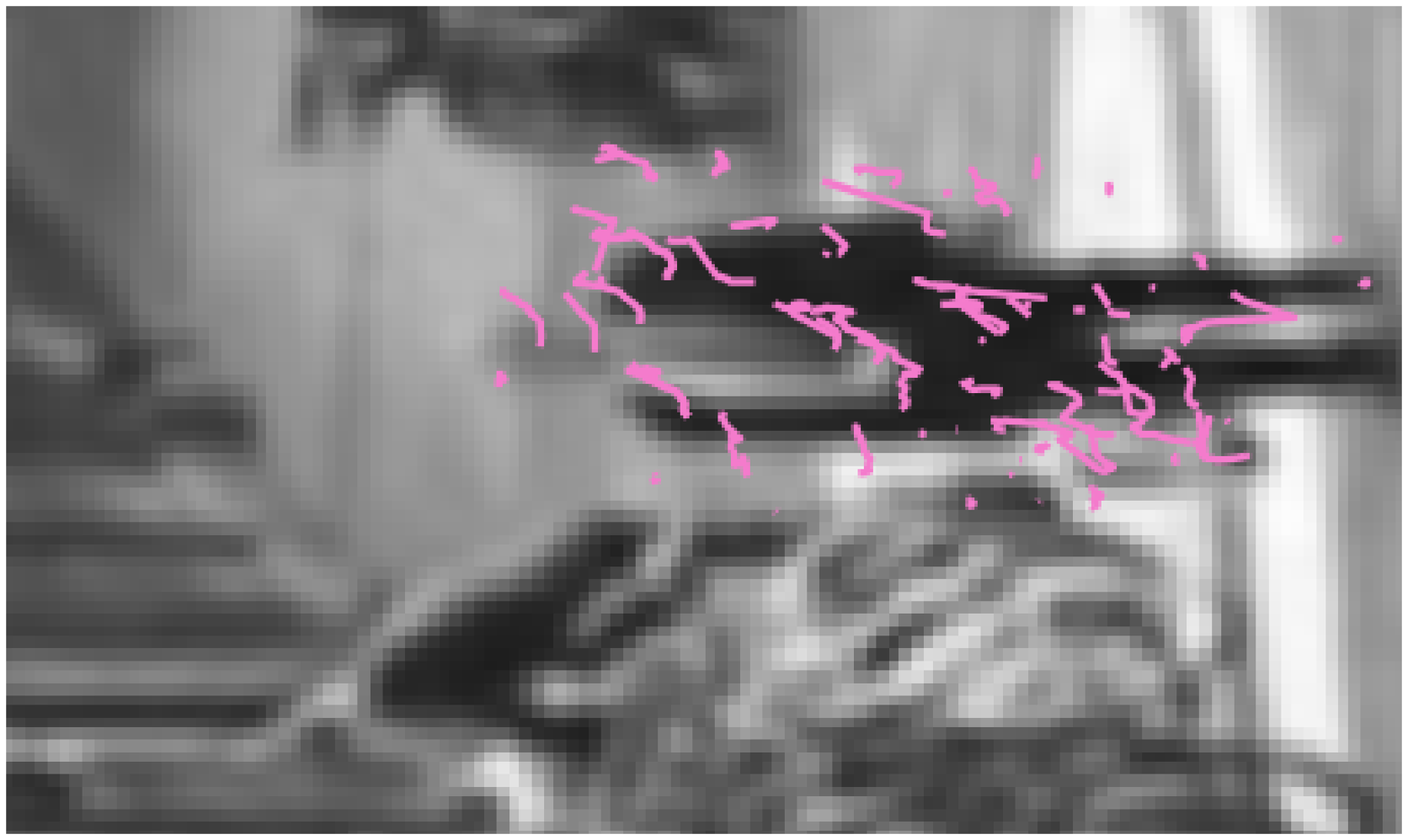}}
    &
    \rotatebox{-90}{\includegraphics[width=0.4\textwidth]{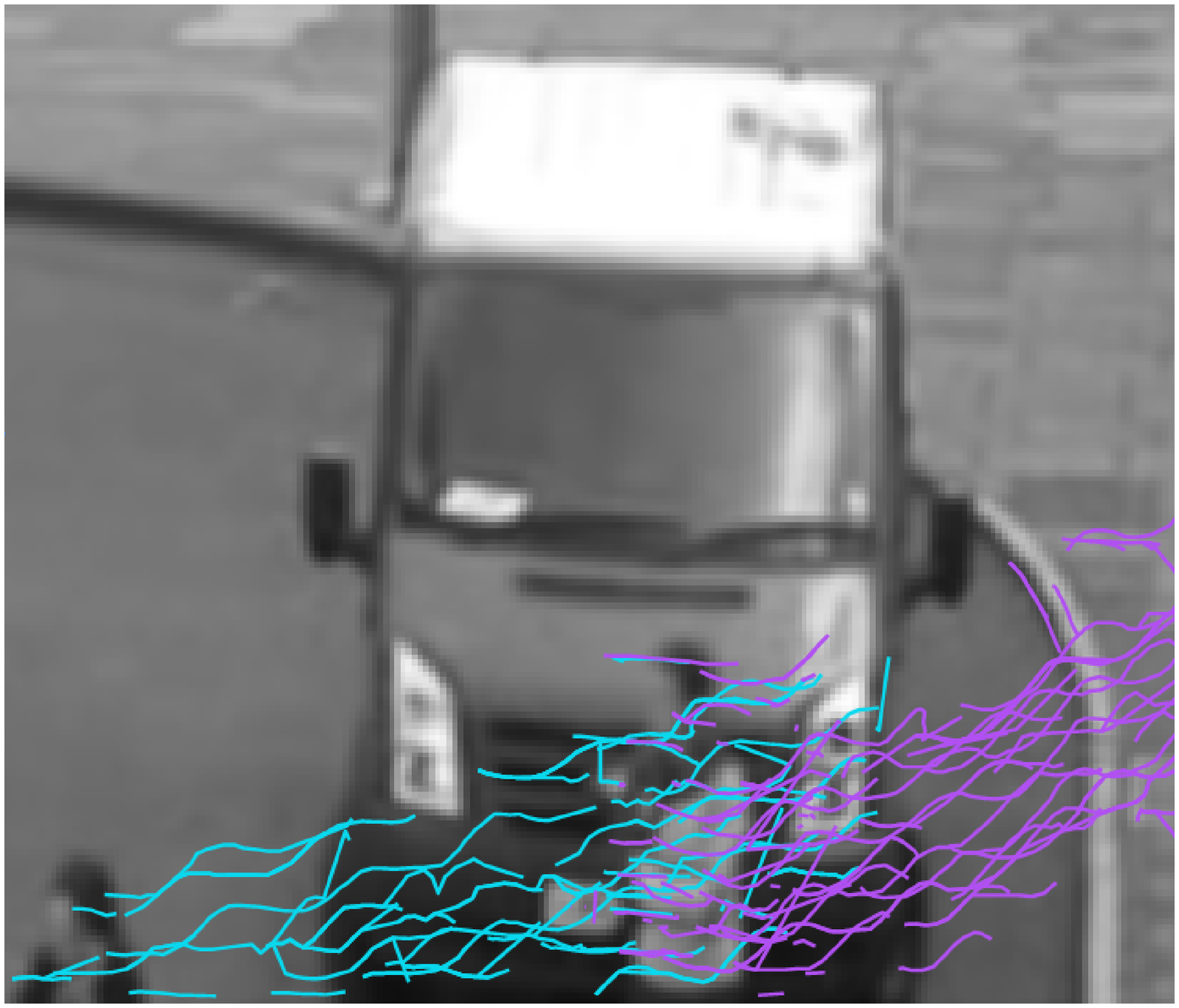}}\\&\\
    (c) & (d)\\&\\
  \end{tabular}
  \caption{Experimental results on a nearly two hour long video footage of a busy marketplace
           confirm the advantages of our method predicted from theory. Feature tracks corresponding
           to each person's model are shown in different colours. Illustrated is our method's
           robustness to (a) mutual occlusion
           of tracked objects, (b,c) successful tracking of an object in the presence of
           scale change, and unstable and changing set of detected local features associated with
           the object's appearance model,
           and (d) a combination of mutual occlusion, cluttered and texture-rich background, scale
           change and gradual disappearance of a tracked object from the scene.}
  \label{f:tracks}
\end{figure}

Experimental results we obtained corroborate previously stated
strengths of our method expected from theory. The permanence of the
proposed model which captures all seen appearances of a tracked
object, coupled with a robust frame-to-frame feature matching, makes
it particularly resilient to errors caused by occlusion. An example
of this can be seen in Figure~\ref{f:tracks}~(a). It shows feature
tracks associated with automatically learnt appearance models
corresponding to two people (shown in different colours -- green and
purple), which are then successfully tracked even following their
mutual occlusion, that is, after one passes in front of the other.

A magnification of a isolated person being tracked in
Figure~\ref{f:tracks}~(b) and another at an approximate 50\% smaller
scale in Figure~\ref{f:tracks}~(c), serve to illustrate the role of
several building elements of our algorithm. Specifically, it can be
observed that few features last for more than 0.5s in the former
example and more than 0.1s in the latter. This is a consequence of
appearance change due to motion and articulation, as well as image
and spatial discretization noise. It is the incremental nature of
our algorithm, whereby novel features are added to the existing
model, and the use of spatio-kinetic clusters, which allows all of
the shown tracks to be associated with the same moving object. These
examples should not be correctly tracked by such previously proposed
method as those of Rabaud and Belongie \cite{RabaBelo2006}, and
Brostow and Cipolla\cite{BrosCipo2006}.

Finally, Figure~\ref{f:tracks}~(d) shows successful tracking in the
presence of several simultaneous difficulties: the two tracked
people cross paths, mutually occluding, in front of a feature-rich
object, one of them progressively disappearing from the scene and
both of them changing in scale due to the movement direction. As
before, many of the associated features are short lived,
disappearing and re-appearing erratically.

\section{Summary and Conclusions}
In this paper we described a novel method capable of automatically
detecting moving objects in complex cluttered scenes, building their
appearance models and tracking them in the presence of partial and
full occlusions, change in appearance (e.g.\ due to articulation or
pose changes) and scale. The proposed algorithm was empirically
evaluated on a two hour long video footage of a busy city
marketplace and the claimed theoretical properties of the approach
substantiated by through successful performance on several difficult
examples involving the aforementioned challenges.

\bibliographystyle{splncs}
\bibliography{my_bibliography}

\begin{thebibliography}{10}

\bibitem{WrenAzarDarrPent1997}
Wren, C., Azarbayejani, A., Darrell, T., Pentland, A.:
\newblock Pfinder:real-time tracking of the human body.
\newblock IEEE Transactions on Pattern Analysis and Machine Intelligence
  (TPAMI) \textbf{19} (1997)  780--785

\bibitem{HariHarwDavi2000}
Haritaoglu, I., Harwood, D., David, L.:
\newblock W4:real-time surveillance of people and their activities.
\newblock IEEE Transactions on Pattern Analysis and Machine Intelligence
  (TPAMI) \textbf{22} (2000)  809--830

\bibitem{ZhaoNeva2004}
Zhao, T., Nevatia, R.:
\newblock Tracking multiple humans in crowded environment.
\newblock In Proc. IEEE Conference on Computer Vision and Pattern Recognition
  (CVPR) \textbf{2} (2004)  406--413

\bibitem{IsarMacC2001}
Isard, M., MacCormick, J.:
\newblock Bramble: a {B}ayesian multiple-blob tracker.
\newblock In Proc. IEEE Conference on Computer Vision and Pattern Recognition
  (CVPR) \textbf{2} (2001)  34--41

\bibitem{RabaBelo2006}
Rabaud, V., Belongie, S.:
\newblock Counting crowded moving objects.
\newblock In Proc. IEEE Conference on Computer Vision and Pattern Recognition
  (CVPR) (2006)

\bibitem{BrosCipo2006}
Brostow, G.J., Cipolla, R.:
\newblock Unsupervised {B}ayesian detection of independent motion in crowds.
\newblock In Proc. IEEE Conference on Computer Vision and Pattern Recognition
  (CVPR) \textbf{1} (2006)  594--601

\bibitem{Gavr2000}
Gavrila, D.M.:
\newblock Pedestrian detection from a moving vehicle.
\newblock In Proc. European Conference on Computer Vision (ECCV) \textbf{2}
  (2000)  37--49

\bibitem{ViolJoneSnow2003}
Viola, P., Jones, M., Snow, D.:
\newblock Detecting pedestrians using patterns of motion appearance.
\newblock In Proc. IEEE International Conference on Computer Vision (ICCV)
  (2003)  734--741

\bibitem{TuSebaDoreKrah2008}
Tu, P., Sebastian, T., Doretto, G., Krahnstoever, N., Rittscher, J., Yu, T.:
\newblock Unified crowd segmentation.
\newblock In Proc. European Conference on Computer Vision (ECCV) \textbf{4}
  (2008)  691--704

\bibitem{ZhaoNevaLv2001}
Zhao, T., Nevatia, R., Lv, F.:
\newblock Segmentation and tracking of multiple humans in complex situations.
\newblock In Proc. IEEE Conference on Computer Vision and Pattern Recognition
  (CVPR) \textbf{2} (2001)  194--201

\bibitem{LiptFujiSapi1998}
Lipton, A., Fujiyoshi, H., Patil, R.:
\newblock Moving target classification and tracking from real-time video.
\newblock In Proc. DARPA Image Understanding Workshop (IUW) (1998)  8--14

\bibitem{MattIshiBake2004}
Matthews, I., Ishikawa, T., Baker, S.:
\newblock The template update problem.
\newblock IEEE Transactions on Pattern Analysis and Machine Intelligence
  (TPAMI) \textbf{26} (2004)  810--815

\bibitem{Lowe2001}
Lowe, D.G.:
\newblock Local feature view clustering for {3D} object recognition.
\newblock In Proc. IEEE Conference on Computer Vision and Pattern Recognition
  (CVPR) (2001)  682--688

\end{thebibliography}
\end{document}